\documentclass[11pt, letterpaper]{arxiv}

\usepackage[all]{hypcap}
\usepackage[comma,numbers,sort,compress]{natbib}
\usepackage[capitalize,noabbrev]{cleveref}
\usepackage{graphicx}
\usepackage{subfigure}
\usepackage{float}
\usepackage{wrapfig}
\usepackage{placeins}

\usepackage{booktabs}
\usepackage{colortbl}
\usepackage{threeparttable}
\usepackage{multirow}
\usepackage{tabularx}
\usepackage{fleqn, tabularx}

\usepackage{amsmath}
\usepackage{amssymb}
\usepackage{mathtools}
\usepackage{amsthm}

\usepackage{algorithm}
\usepackage{algpseudocode}

\usepackage{microtype}
\usepackage{setspace}
\usepackage{xcolor}
\usepackage[most]{tcolorbox}

\usepackage{enumitem}

\usepackage{listings}
\usepackage{color}

\definecolor{sb_blue}{RGB}{65,105,225}
\definecolor{sb_orange}{RGB}{255,140,0}
\definecolor{gray}{RGB}{128,128,128}

\lstdefinestyle{mystyle}{
    language=Python,
    xleftmargin=5.0ex,
    basicstyle=\footnotesize\ttfamily\linespread{4},
    backgroundcolor=\color{gray!10},
    commentstyle=\color{gray},
    alsoletter={<>-0123456789},
    keywordstyle=\color{sb_blue},
    ndkeywords={nn, F, torch, partial},
    ndkeywordstyle=\color{sb_orange},
    emph={import, def, return, if, else},
    emphstyle=\bfseries\color{sb_blue},
    numberstyle=\footnotesize\ttfamily\color{gray},
    stringstyle=\color{sb_blue},
    breakatwhitespace=false,
    breaklines=true,
    keepspaces=true,
    numbers=left,
    numbersep=5pt,
    showspaces=false,
    showstringspaces=false,
    showtabs=false,
    tabsize=2
}
\definecolor{lightblue}{rgb}{0.22,0.45,0.70}
\newcommand{\Appendix}{\textcolor{mylinkcolor}{Appendix}}
\newcommand{\Figure}{\textcolor{mylinkcolor}{Figure}}
\newcommand{\Section}{\textcolor{mylinkcolor}{Section}}
\newcommand{\Table}{\textcolor{mylinkcolor}{Table}}

\title{Beyond One-Size-Fits-All: Diagnosis-Driven Online Reinforcement Learning with Offline Priors}

\author[1]{Guozheng Ma}
\author[2]{Lu Li}
\author[3]{Zilin Wang}
\author[2]{Pierre-Luc Bacon}
\author[1]{Dacheng Tao}

\affil[1]{Nanyang Technical University}
\affil[2]{Mila - Quebec AI Institute $\&$ Université de Montréal}
\affil[3]{University of Oxford}

\begin{abstract}

Online reinforcement learning (RL) agents increasingly depend on knowledge acquired offline to achieve practical efficiency. Originally studied in offline-to-online RL, this paradigm now spans foundation model post-training and embodied intelligence, with prior types expanding from offline datasets and pre-trained policies to increasingly diverse knowledge sources such as multimodal foundation models and generative world models. Offline priors have become central to how deep RL is developed and deployed. However, this reliance introduces a challenge that the prevailing benchmark-driven paradigm cannot resolve: because prior validity varies across deployments and shifts during training, no single approach to managing it is universally optimal, and benchmark rankings offer limited guidance for real-world deployments. Rather than pursuing universal solutions, we argue that the field should shift to diagnosis-driven tension management, in which deployment-specific evidence guides how the learner relates to its priors throughout training, enabling both flexible and adaptive deployment. We support this position with a framework characterizing how priors reshape online optimization through three functional roles, controlled experiments demonstrating help-or-hurt reversals, cross-domain evidence from foundation model post-training to embodied intelligence, and engagement with five substantive counterarguments.   % Abstract content in a separate file
\end{abstract}

\begin{document}
\maketitle

% ===================== 定制化目录开始 =====================
\vspace{-\baselineskip}
\begingroup
    % 1. 隐藏默认的 "Contents" 标题
    \makeatletter
    \renewcommand{\tableofcontents}{\@starttoc{toc}}
    \makeatother

    % 2. 局部修改超链接颜色为主题蓝 (基于你模板中的 BerkeleyBlue)
    \hypersetup{linkcolor=BerkeleyBlue}

    % 3. 精细控制目录的行间距与字体样式 (使用你模板已包含的 titletoc 宏包)
    % 相比全局修改 linespread，这种方法只控制 section 条目之间的距离，是最优雅的做法
    \titlecontents{section}[0em]
        {\vspace{4pt}\large\bfseries}
        {\thecontentslabel\quad}       % 序号和标题之间的间距
        {}                             % 无序号标题的格式
        {\hfill\contentspage}          % 页码靠右对齐

    % 4. 生成一级目录
    \setcounter{tocdepth}{1}
    \tableofcontents
\endgroup % 在目录和引言 (Introduction) 之间留出适当的空白间距
% ===================== 定制化目录结束 =====================

\newpage
\section{Introduction}
\label{sec:intro}

Online reinforcement learning (RL) agents increasingly rely on knowledge acquired offline to achieve practical performance.
In foundation model post-training, large-scale pre-trained models provide the base capability that online RL refines for alignment, reasoning, and agentic applications~\citep{ouyang2022training, guo2025deepseek}.
In embodied intelligence, simulators, demonstration datasets, world models, and pre-trained policies supply the prior knowledge that makes online learning on physical platforms feasible~\citep{peng2018sim, hafner2023mastering, walke2023bridgedata, luo2024serl}.
Despite differences in domain, objective, and prior type, offline priors now provide the foundation from which agents learn through online experience~\citep{ball2023efficient, zhou2025efficient, silver2025welcome, lewandowski2026the}.

%%%%%%%%%%%%%%%%%%%%%%%%% Paragraph 2 %%%%%%%%%%%%%%%%%%%%%%%%%
Despite these benefits, how to use offline priors remains a persistent challenge: the same reliance decision that helps in one setting often hurts in another.
In offline-to-online RL, whether to preserve the pre-trained policy, retain the offline dataset, or maintain conservative value estimates each produces different outcomes depending on the quality of the offline sources and their relationship to the deployment task~\cite{li2025three, zhou2025efficient, ball2023efficient, wang2023train}.
In foundation model post-training, the role of the reference constraint varies with the fidelity of the reward signal: when rewards are verifiable, strong regularization restricts the discovery of novel strategies~\cite{yu2025dapo}, while when rewards come from learned preference models, the same regularization is essential for preventing overoptimization of proxy scores~\cite{ouyang2022training, gao2023scaling}.
In robotics, whether to train in simulation and transfer or to learn directly on physical hardware produces different outcomes depending on the fidelity of the simulator and the complexity of the contact dynamics~\citep{as2026matters, luo2024serl, levy2026simulation}.
These inconsistencies recur across communities, domains, and prior types, suggesting a shared structural property rather than isolated engineering issues.

%%%%%%%%%%%%%%%%%%%%%%%%% Paragraph 3 %%%%%%%%%%%%%%%%%%%%%%%%%
We argue that these inconsistencies reflect an inherent conflict: offline priors carry knowledge that is inevitably bounded, while online RL exists precisely to push beyond those bounds.
\textcolor{mydarkgreen}{$\bullet$}~Specifically, whether the prior takes the form of a pre-trained policy, a simulator, or an offline dataset, it encodes knowledge acquired under conditions that may differ from those of deployment. As a result, the extent to which this knowledge remains useful is uncertain and can only be revealed through online interaction.
We call this the \textit{bounded commitment} of offline priors: valuable knowledge with an inherently uncertain scope of validity.
\textcolor{mydarkgreen}{$\bullet$}~On the other hand, online RL drives the agent toward optimal performance through interaction with the deployment environment.
This requires the agent to eventually surpass what the prior covers, while relying on its knowledge to get there efficiently.
As long as the agent operates within the scope of the prior's validity, reliance is purely beneficial. Once learning pushes beyond that scope, a genuine tension emerges: stronger reliance constrains adaptation, while freer adaptation risks discarding knowledge that still helps elsewhere.
Crucially, the agent cannot know where this boundary lies, and the boundary itself shifts as online experience accumulates.
Unlike a static trade-off that can be settled at design time, this tension persists and evolves throughout learning. Hence, tension management is a central challenge of this paradigm.

% Since the right reliance varies with the task, the environment, and the quality of the prior, no single configuration is universally optimal.

%%%%%%%%%%%%%%%%%%%%%%%%% Paragraph 4 %%%%%%%%%%%%%%%%%%%%%%%%%
One consequence is that tension management has no universal optimum: the right reliance on each prior depends on the deployment and shifts as learning progresses.
Despite this, the field continues to evaluate progress by comparing methods on fixed benchmarks, implicitly assuming that the resulting rankings reflect universal truths rather than deployment-specific matches.
The field accumulates condition-specific performance rankings rather than transferable understanding.
The way forward requires a shift in research perspective. Rather than asking which method is best, the field should ask what determines when each design choice helps or hurts.

\begin{figure}[t]
\centering
\vspace{-\baselineskip}
\includegraphics[width=\textwidth]{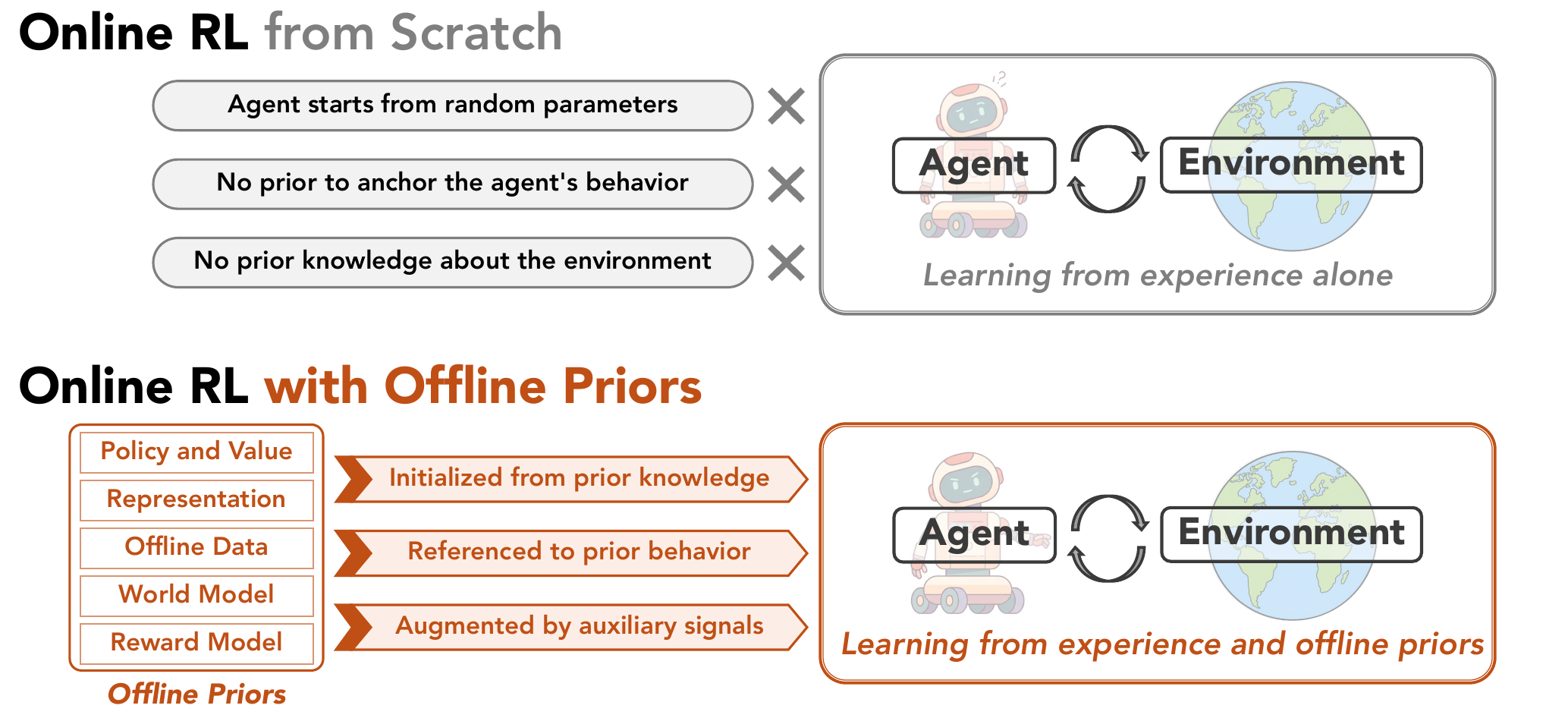}
\vspace{-1.5\baselineskip}
\caption{Online RL from scratch (top) versus with offline priors (bottom). The agent-environment interaction loop is identical in both cases. Offline priors add knowledge sources that accelerate learning but also introduce structural tensions analyzed in \Section~\ref{sec:tensions}.
}
\vspace{-1\baselineskip}
\label{fig:paradigm}
\end{figure}

%%%%%%%%%%%%%%%%%%%%%%%%% Position statement %%%%%%%%%%%%%%%%%%%%%%%%%
\state{We argue that the field of online RL with offline priors should move beyond one-size-fits-all methods toward diagnosis-driven tension management. Since no universal optimum exists for how agents should rely on their priors, effective deployment requires diagnostic infrastructure that can assess the prior-deployment match and monitor how tensions evolve during learning, enabling both flexible and adaptive deployment.}

%%%%%%%%%%%%%%%%%%%%%%%%% Paper structure %%%%%%%%%%%%%%%%%%%%%%%%%
We develop this position as follows. \Section~\ref{sec:paradigm} defines the paradigm and introduces a taxonomy of offline prior types. \Section~\ref{sec:tensions} analyzes why offline priors sharpen the core tensions of online RL and introduces the concept of bounded commitments. \Section~\ref{sec:diagnostics} presents evidence that tension management has no universal optimum and argues for a shift from benchmark-driven to diagnosis-driven tension management. \Section~\ref{sec:alternative} engages with several potential objections and counterarguments to our core position. \Section~\ref{sec:conclusion} concludes with research opportunities and a broader perspective.          % Introduction
% \newpage
\section{The Paradigm of Online RL with Offline Priors}
\label{sec:paradigm}
 
Instead of \textit{\textbf{learning from scratch}},
% \footnote{Also referred to as \textit{tabula rasa} RL~\cite{agarwal2022reincarnating}.}, 
online RL agents increasingly learn not only from their own online interaction with the environment but also from knowledge acquired offline: setting initial parameters, constraining how far the learner may deviate from prior behavior, or supplying supplementary data or predictions alongside real interaction.
Though studied under different names across offline-to-online RL~\cite{nair2020awac, nakamoto2023cal}, LLM post-training~\cite{ouyang2022training, guo2025deepseek}, sim-to-real transfer~\cite{wagenmaker2024overcoming, as2026matters}, model-based RL~\cite{hafner2023mastering, hansen2024tdmpc}, and vision-language-action model fine-tuning~\cite{zitkovich2023rt, guo2025improving}, these mechanisms define a common paradigm that we call \textbf{\textit{online RL with offline priors}}. 
As in any RL system, the interaction loop has two sides: an agent that selects actions and an environment that produces states and rewards.
Offline knowledge can concern either side, and we organize the resulting priors accordingly.

% Offline priors naturally fall along the fundamental boundary of the RL loop: priors about the agent (policy, value, representation) encode how to act, while priors about the environment (dynamics, reward, data) encode how the world behaves.

\begin{table}[ht]
\centering
\caption{Taxonomy of offline priors in online RL, classified by \textit{what knowledge they encode} and \textit{how they function} during online learning. Agent-side priors encode knowledge about how to act or evaluate; environment-side priors encode knowledge about how the world behaves.}
\label{tab:prior-taxonomy}
\small
\begin{threeparttable}
\begin{tabular}{@{}cc ccc @{}}
\toprule
\textbf{Knowledge Side} & \textbf{Functional Role} & \multicolumn{3}{c}{\textbf{Representative Prior Type}} \\
\midrule
\multirow{2}{*}{Agent} 
  & Initialization & Policy $\pi_0$ & Value $Q_0$ or $V_0$ & Representation $\phi_0$ \\
\cmidrule(l){2-5}
  & Reference      & Policy $\pi_{\mathrm{ref}}$ & Value $Q_{\mathrm{ref}}$ or $V_{\mathrm{ref}}$ & \tnote{\dag}~~Offline Data $\mathcal{D}$ \\
\midrule
Environment
  & Auxiliary       & \tnote{*}~~World Model $\hat{M}$ & Reward Model $\hat{R}$ & \tnote{\dag}~~Offline Data $\mathcal{D}$\\
\bottomrule
\end{tabular}
\begin{tablenotes}
\footnotesize
\item[\dag] Offline data $\mathcal{D}$ records how the environment transitions and how the collecting policy acts; its functional role depends on which aspect the online algorithm extracts~\cite{ball2023efficient, fujimoto2021minimalist}.
\item[*] World model $\hat{M}$ spans a broad spectrum: compact latent dynamics models~\cite{hafner2023mastering, hansen2024tdmpc}, engineered simulators used to supplement real-world interaction~\cite{wagenmaker2024overcoming, as2026matters}, and world foundation models~\cite{hou2026worldmodelrobotlearning, agarwal2025cosmos, team2026gigabrain, ye2026worldactionmodelszeroshot}.
\end{tablenotes}
\end{threeparttable}
\end{table}
% How a prior is used matters as much as what it encodes.
Offline priors can shape online optimization through exactly three channels: by determining where optimization begins, by constraining what objective it pursues, or by providing experience beyond direct interaction.
\textcolor{mydarkgreen}{$\bullet$}~Initialization priors set the starting point for online learning: an offline-trained policy in offline-to-online RL~\cite{nakamoto2023cal}, a supervised fine-tuned model in LLM post-training~\cite{ouyang2022training}, or a simulation-trained controller in sim-to-real transfer~\cite{peng2018sim} each play this role, providing initial competence and reducing the exploration burden that dominates learning from scratch. 
\textcolor{mydarkgreen}{$\bullet$}~Reference priors modify the learning objective by anchoring updates to prior behavior: the KL penalty to a reference policy in RLHF~\cite{ouyang2022training}, the conservative value penalty in Cal-QL~\cite{nakamoto2023cal}, and behavioral cloning regularization~\cite{fujimoto2021minimalist, lu2023imitation} are different mechanisms serving the same structural function. In each case, the prior defines a trust region that the online learner is penalized from leaving.
\textcolor{mydarkgreen}{$\bullet$}~Auxiliary priors provide additional information outside the online loop, whether through retained offline data~\cite{ball2023efficient}, model-based rollouts~\cite{hafner2023mastering}, or learned reward signals~\cite{ouyang2022training}. Unlike reference priors, these do not alter what the optimizer aims to achieve, but expand the evidence it can draw on.

Offline data occupies a unique position in this taxonomy.
It may originate from behavior-policy rollouts, expert demonstrations, human preference comparisons, or in-the-wild recordings~\cite{shaw2022videodex, walke2023bridgedata}, and it inherently records both how the environment transitions and how the collecting policy acts.
Which aspect the online algorithm extracts determines the functional role: replay for value estimation treats data as auxiliary information~\cite{ball2023efficient}, while behavioral regularization treats the same data as a reference~\cite{fujimoto2021minimalist, wu2019behavior}.
This dependence on algorithmic use extends beyond data: a pessimistic value function always initializes the learner~\cite{nakamoto2023cal, zhou2025efficient}, but additionally serves as an ongoing reference when its conservative penalty is maintained during fine-tuning~\cite{kumar2020conservative}.
In general, functional role is not an inherent property of any prior; it is determined by the algorithm that deploys it.

These priors fundamentally reshape the online learning process.
The next section examines why these changes, despite their well-documented benefits, introduce structural tensions into online learning.    % Preliminaries / Background
% \newpage
\section{Why Offline Priors Sharpen Tensions}
\label{sec:tensions}

% Offline priors are designed to address the core difficulties of online RL. 
In principle, two fundamental tensions govern the online RL learning process: exploration must be balanced against exploitation, and the plasticity to incorporate new experience must be balanced against the stability of what has already been learned. However, in the challenging tasks that organize mainstream deep RL research the balance tilts sharply to one side.
Similarly, sparse rewards, high-dimensional action spaces, and long horizons make useful discoveries so rare that the need for exploration overwhelms any concern about premature exploitation~\cite{ecoffet2021first, ladosz2022exploration}.
Bootstrapping from the agent's own shifting value estimates induces optimization pathologies that progressively degrade the network's ability to incorporate new experience~\cite{nikishin2022primacy, lyle2023understanding, dohare2024loss, ma2024revisiting}.
In these settings, failure does not stems from unbalanced tensions, but typically from one-side insufficiency: agents explore too little to discover useful behavior and remain too rigid to learn effectively from new evidence.
The research priorities of the past decade reflect this asymmetry: exploration methods overwhelmingly aim to increase coverage~\cite{hao2023exploration, ladosz2022exploration}, and plasticity interventions overwhelmingly aim to restore adaptability~\cite{nikishin2023deep, sokar2023dormant, klein2024plasticity}.

\textbf{How Priors Sharpen the Tensions.}~
Over the past several years, diverse research communities have effectively addressed these bottlenecks by equipping agents with offline priors before online interaction begins. Pre-trained policies and value functions reduce the exploration burden by providing informed starting behavior rather than random search~\cite{nair2020awac, nakamoto2023cal, ouyang2022training}. Offline data and world models ground the learning process in prior experience, mitigating the cold-start pathologies that degrade network capacity from the earliest updates~\cite{ball2023efficient, hafner2023mastering}. These gains are substantial and well documented.
However, as priors grow stronger, the previously negligible side of each tension becomes increasingly consequential, and the full two-sided character of both oppositions re-emerges.

This shift is visible along both axes.
\textcolor{mydarkgreen}{$\bullet$}~On the \textit{stability-plasticity} axis, aggressive online updates risk catastrophic forgetting at the offline-to-online transition~\cite{zhou2025efficient, luo2023finetuning, wolczyk2024finetuning}, while distributional mismatch between offline data and online rollouts can destabilize value estimation~\cite{ball2023efficient}.
Meanwhile, plasticity failure is no longer only a matter of network capacity, such as dormant neurons or rank collapse~\cite{sokar2023dormant, dohare2024loss}, but also one of optimization bias, where strong initialization shapes the loss landscape in ways that bias subsequent training toward the prior~\cite{li2025three, lyle2024disentangling}.
\textcolor{mydarkgreen}{$\bullet$}~On the \textit{exploration-exploitation} axis, strong priors can narrow the agent's policy toward pre-trained behavioral modes~\cite{zhao2025echo, yue2025does}, while optimization against learned reward models can drive the agent to exploit proxy scores rather than explore genuinely better behavior~\citep{gao2023scaling}.
Meanwhile, exploration failure shifts from an inability to reach informative states ~\cite{ecoffet2021first} to a difficulty in moving beyond the signals carried by the prior~\cite{zhao2025echo}.
In both cases, priors do not simply solve the bottlenecks; they restore and sharpen the full two-sided character of a tension that task difficulty had compressed into a one-sided bottleneck.

\textbf{Priors as Bounded Commitments.}~
The sharpened tensions described above share a common structural root.
Every offline prior encodes knowledge from a source setting that may differ from the deployment environment. 
The agent cannot fully determine where this knowledge remains valid and where it does not, yet it must \textit{rely on} the prior to learn efficiently and \textit{adapt beyond} it where the prior falls short.
We call this epistemic status a \textbf{\textit{bounded commitment}}: the prior is valuable but its scope of validity is bounded, and the agent must commit to using it without knowing those bounds precisely.
The insight that source knowledge has limited validity in new contexts is well established across Bayesian RL~\cite{ghosh2022offline, hu2024bayesian}, transfer learning~\cite{zhang2022survey}, and adaptive offline RL~\cite{ni2026adaptive}. The concept of `bounded commitments’ is precisely a name given to this fundamental insight, as it applies to all previous types of online RL.
\Table~\ref{tab:structural} makes this concrete: each functional role from \Section~\ref{sec:paradigm} introduces a reliance parameter ($\mu$, $\lambda$, $\beta$) that governs how strongly the learner commits to its priors.

\begin{table}[ht]
\centering
\caption{Structural changes that offline priors introduce to online RL optimization. The reliance parameters $\mu$, $\lambda$, $\beta$ each govern how strongly the learner relies on the prior; setting any to zero recovers the from-scratch case. $\mathcal{M}_{\mathrm{prior}}$ denotes the effective (possibly approximate) environment implied by auxiliary sources ($\mathcal{D}$, $\hat{M}$, $\hat{R}$), and $J(\theta)$ is shorthand for $J(\theta;\,\mathcal{M}_{\mathrm{deploy}})$.}
\label{tab:structural}
\small
\begingroup
\setlength{\aboverulesep}{0pt}
\setlength{\belowrulesep}{0pt}
\resizebox{\textwidth}{!}{%
\begin{tabular}{@{}lc>{\columncolor{mycitecolor!8}}cc@{}}
\toprule
\rule[-0.8ex]{0pt}{3.0ex}
 & \textbf{Online RL from Scratch}
 & \textbf{Online RL with Offline Priors}
 & \textbf{Reliance} \\
\midrule
\rule{0pt}{3ex}%
\textbf{(a) Initialization}
& $\theta_0 \sim \mathrm{random}$ 
& $\theta_0 = \mu\,\theta_{\mathrm{prior}} + (1\!-\!\mu)\,\theta_{\mathrm{rand}}$ 
& $\mu \in [0,1]$ \\[8pt]
\textbf{(b) Reference}
& $\displaystyle\max_\theta\; J(\theta)$ 
& $\displaystyle\max_\theta\; J(\theta) - \lambda\, L_{\mathrm{ref}}(\theta)$ 
& $\lambda \geq 0$ \\[8pt]
\textbf{(c) Auxiliary}
& $\displaystyle\max_\theta\; J(\theta;\, \mathcal{M}_{\mathrm{deploy}})$ 
& $\displaystyle\max_\theta\; (1\!-\!\beta)\,J(\theta;\, \mathcal{M}_{\mathrm{deploy}}) + \beta\,J(\theta;\, \mathcal{M}_{\mathrm{prior}})$ 
& $\beta \in [0,1]$ \\[2pt]
\bottomrule
\end{tabular}%
}
\endgroup
\end{table}

Because the validity boundary of each prior is never fully knowable, every reliance configuration is necessarily a bet. Although \Table~\ref{tab:structural} expresses reliance through continuous parameters for analytical clarity, $\mu$, $\lambda$, and $\beta$ abstract over method-level design decisions such as whether to use pre-trained weights, whether to constrain the objective, and whether to supplement the replay buffer. Too much reliance risks trapping the agent in knowledge that does not hold; too little wastes knowledge that could have accelerated learning. This difficulty is structural, not algorithmic, and is compounded by coupling: the KL penalty $\lambda$ in RLHF, for instance, simultaneously controls policy stability and exploration freedom~\citep{zhao2025echo}, so that adjusting one tension inevitably affects the other. Because the right reliance depends on how well each prior matches the deployment environment, \Section~\ref{sec:diagnostics} examines how this challenge manifests empirically and what it implies for how the field evaluates progress.

\Takeaway{Offline priors are fundamentally bounded commitments: valuable knowledge whose scope of validity the agent can never fully determine. This is why offline priors, despite their substantial benefits, sharpen rather than resolve the core tensions of online RL.}

% \newpage
\section{From Pursuing Universality to Diagnosis-Driven Flexibility and Adaptivity}
\label{sec:diagnostics}

Despite the structural tensions identified in \Section~\ref{sec:tensions}, the current dominant research paradigm continues to evaluate progress by comparing methods on fixed benchmarks and seeking algorithms that perform well across the board.
This approach implicitly assumes that a single reliance configuration can be universally optimal. 
This section presents evidence against that assumption (\Section~\ref{sec:no-universal}) and argues that reliable deployment requires diagnostic infrastructure rather than universal methods (\Section~\ref{sec:call}).

\begin{figure}[t]
\centering
% \vspace{-\baselineskip}
\includegraphics[width=\textwidth]{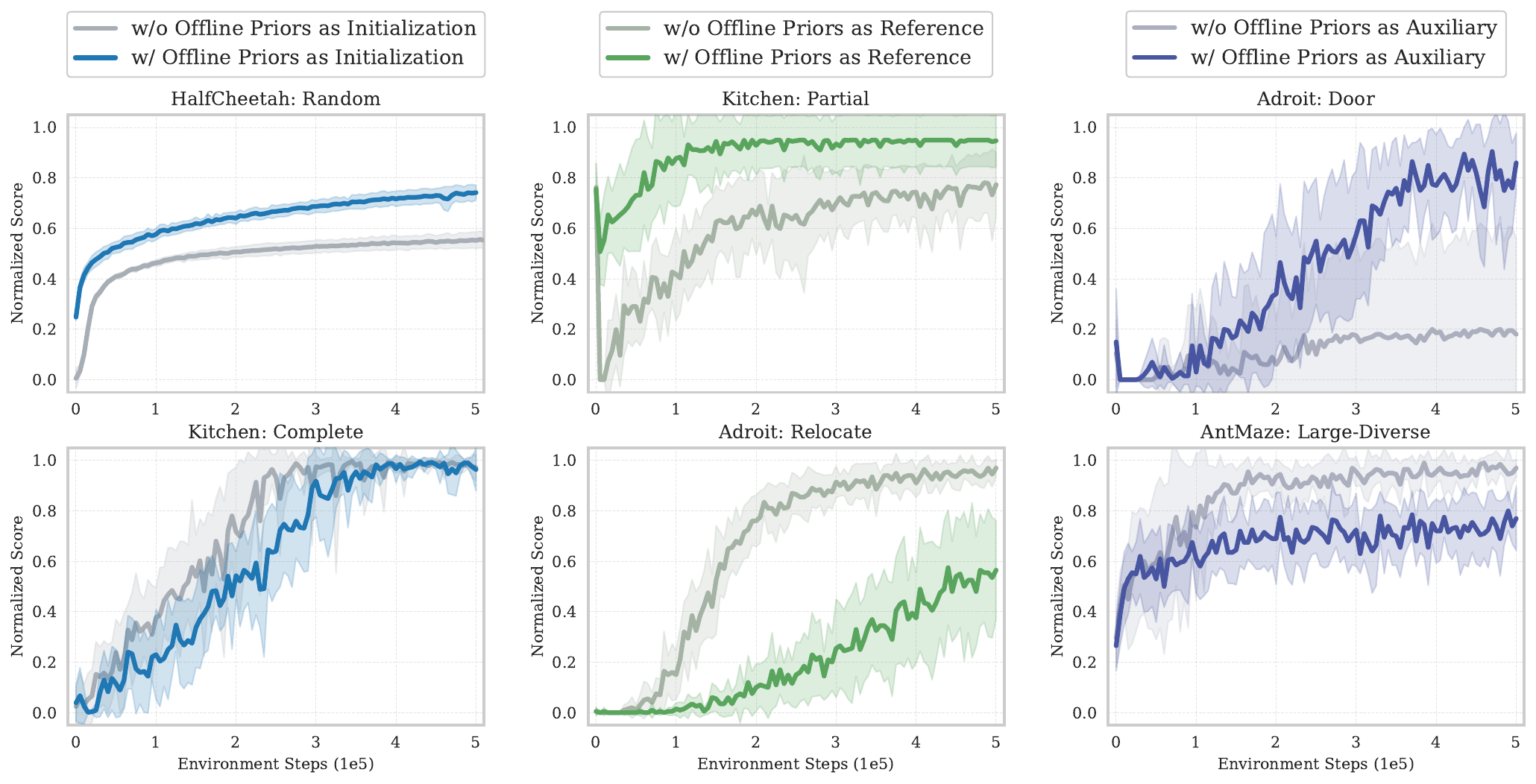}
% \vspace{-2\baselineskip}
\caption{No single reliance configuration is universally optimal. Each column isolates one reliance parameter from \Table~\ref{tab:structural}: initialization~($\mu$, left), reference~($\lambda$, middle), and auxiliary~($\beta$, right). The top row shows a task where stronger reliance on the prior helps; the bottom row shows a task where the same choice hurts. \textbf{Left}: RLPD with versus without offline-trained initialization. \textbf{Middle}: Cal-QL's conservative penalty versus unconstrained updates, with both initialization and offline data present. \textbf{Right}: retaining versus discarding offline data, with both initialization and conservative penalty present. Full results and additional controlled comparisons are provided in \Appendix~\ref{app:experiments}.}
% \vspace{-1.5\baselineskip}
\label{fig:reversal}
\end{figure}

\subsection{Tension Management Has No Universal Optimum}
\label{sec:no-universal}

How well each prior matches the deployment environment determines the optimal reliance configuration: how closely the offline data covers the online distribution, how accurately a world model reflects real dynamics, and how faithfully a pre-trained policy captures the behavior the task requires. 
Since this match varies across tasks and conditions, the optimum varies with it.

\textbf{Illustrative Experiments.}~
We demonstrate this non-universality through controlled experiments in the offline-to-online RL setting. We choose this setting because it abstracts away domain-specific engineering details present in LLM post-training, sim-to-real transfer, and VLA fine-tuning, allowing each reliance parameter from \Table~\ref{tab:structural} to be toggled independently.
\textcolor{mydarkgreen}{$\bullet$}~For initialization~($\mu$), we compare starting from the offline-trained policy weights \textit{versus} resetting to a randomly initialized network.
\textcolor{mydarkgreen}{$\bullet$}~For reference~($\lambda$), we compare maintaining conservative value penalties throughout fine-tuning \textit{versus} dropping all conservatism and using unconstrained optimization~\cite{nakamoto2023cal, haarnoja2018soft}.
\textcolor{mydarkgreen}{$\bullet$}~For auxiliary information~($\beta$), we compare retaining the offline dataset in the replay buffer \textit{versus} discarding it and learning from online data only~\cite{ball2023efficient, zhou2025efficient}.

\Figure~\ref{fig:reversal} presents representative task pairs from our experiments (full results across all tasks in \Appendix~\ref{app:experiments}). In each column, the same binary choice produces opposite outcomes across tasks. These reversals are not random variation: they arise because the optimal reliance depends on a complex interaction between the properties of the prior, the structure of the deployment task, and the degree to which they match. Since all three factors vary across settings, no single configuration is reliably beneficial.
Recent work has made this pattern precise in offline-to-online RL by identifying distinct regimes in which the optimal strategy qualitatively flips~\cite{li2025three}. 
Our experiments extend this observation by showing that non-universality spans all three reliance dimensions independently, not only the interaction between initialization and data retention.

\textbf{A Cross-Domain Pattern.}~
The same non-universality appears far beyond offline-to-online RL. 
\textcolor{mydarkgreen}{$\bullet$}~In LLM post-training, the debate over reference regularization illustrates the point directly: removing the KL penalty improves reasoning performance on verifiable tasks~\cite{yu2025dapo}, but the penalty remains essential for preventing reward hacking when rewards come from learned models~\cite{ouyang2022training, gao2023scaling}. 
Recent work shows that gradient regularization can outperform KL penalties entirely in some regimes while failing in others~\cite{ackermann2026gradient}, and that static length penalties help efficiency on easy tasks but hurt accuracy on hard ones~\cite{peng2026think}. 
\textcolor{mydarkgreen}{$\bullet$}~In sim-to-real robotics, a large-scale study across three robot platforms finds that widely used algorithmic defaults can be harmful on physical hardware~\cite{as2026matters}, and that end-to-end policy fine-tuning collapses in real-world deployment even when it succeeds in simulation~\cite{levy2026simulation}.
\textcolor{mydarkgreen}{$\bullet$}~In vision-language-action model fine-tuning, sequential adaptation with LoRA works remarkably well for large pretrained VLAs but collapses when any single ingredient is removed~\cite{hu2026simple}.
\textcolor{mydarkgreen}{$\bullet$}~In model-based RL, explicit conservatism helps on high-coverage datasets but fails on low-quality data, where Bayesian approaches without conservatism perform better~\cite{ni2025long}. 
Across all these settings, the underlying pattern is the same: the optimal reliance level depends not on the method alone but on the properties of the prior, the demands of the deployment task, and how well the two align.
That this appears independently across communities with different methods, vocabularies, and evaluation practices suggests it is structural rather than incidental.

% Recent diagnostic work in offline-to-online RL has made this pattern precise by identifying distinct regimes in which the optimal strategy qualitatively flips~\cite{li2025three}. Our experiments extend this observation by showing that non-universality spans all three reliance dimensions, not only the interaction between initialization and data retention, and that the same structural phenomenon operates across communities that use different terminology to describe it.

\textbf{The Inherent Limits of Benchmark-Driven Evaluation.}~
% P1: Insight + community response + why it fails
The evidence above implies that each method implicitly encodes a particular reliance configuration. 
A benchmark ranking therefore reflects how well that configuration matches the evaluation conditions, not a universal ordering of methods. When the deployment setting changes, the ranking can change with it.
The standard response to this fragility has been to broaden evaluation by testing on more tasks, more environments, and more data conditions.
However, expanding the evaluation suite tends to multiply contradictory findings rather than eliminate them, and aggregate metrics compress these disagreements into a single ranking that explains none of them. 
As the empirical record grows, what accumulates is not transferable understanding but an expanding catalogue of condition-specific performance rankings.

% P2: Related Position
Recent position papers have raised compatible concerns: that rigorous RL benchmarking is prohibitively expensive~\cite{jordan2024position}, that aggregate scores obscure fragile generalization~\cite{chen2025rethinking}, and that standard protocols hide the true cost of hyperparameter selection~\cite{tang2025position} and mask deployment non-stationarity~\cite{mesbahi2025position}. 
The non-universality we identify raises a more fundamental concern: even methodologically sound benchmarking cannot produce transferable conclusions when the optimal configuration is itself deployment-dependent.

% P3: Root cause + bridge 
The fundamental limitation is not the size or quality of any benchmark but the kind of question that benchmark comparisons can answer. Rankings order methods within a fixed setting; they do not reveal which properties of the prior and the deployment task govern whether a design choice helps or hurts. Progress requires a different kind of question: not which method is best, but what each deployment needs. Without infrastructure to answer that question, the field risks an indefinite cycle of benchmark expansion without convergent insight.

\Takeaway{No single reliance configuration is universally optimal: the same design choice that helps in one setting can hurt in another. Benchmark comparisons cannot resolve this because they answer which method wins, not what determines when each choice helps or hurts.}

%%%%%%%%%%%%%%%%%%%%%%%%%%%%%%%%%%%%%%%%%%%%%%%%%%%%%%%%%%%%%%%%%%%%%
\subsection{From Benchmark-Driven to Diagnosis-Driven Tension Management}
\label{sec:call}

More or better benchmarks cannot overcome a limitation inherent in ranking-based evaluation itself. The field needs to change not the evaluation tools but the question those tools are designed to answer. Concretely, we advocate a shift from \textit{benchmark-driven} to \textit{diagnosis-driven} tension management: from letting fixed rankings guide method selection to letting deployment-specific evidence guide how the learner relates to its priors throughout training.

\textbf{The Core Distinction.}~
In the benchmark-driven paradigm, reliance on each prior is configured before deployment based on aggregate evaluation results, and remains fixed or follows a pre-determined schedule once training begins. Online experience serves only to update the policy, even though every transition also carries evidence about the prior itself, including whether value estimates still align with the deployment environment, whether offline data still overlaps with online rollouts, and whether reference behavior still serves the task. This evidence goes unused. The diagnosis-driven paradigm treats it as a first-class signal. The same transitions that update the policy also reveal whether value estimates remain calibrated, whether offline data still provides useful grounding, and whether reference behavior continues to guide learning productively.
This evidence informs reliance decisions throughout training, determining when to trust the prior and when to move beyond it. 
In short, diagnosis-driven tension management is the practice of systematically extracting evidence about prior validity from online interaction and using it to guide reliance decisions.

This is not merely a methodological preference. 
However carefully a prior is constructed, its validity boundary in any specific deployment remains uncertain until interaction begins.
Offline evaluation can estimate prior quality in general, but cannot determine which specific aspects will hold or fail under new conditions.
Designing better priors or more robust algorithms can reduce the frequency of severe mismatches, but cannot eliminate the underlying uncertainty: the agent is always using knowledge acquired elsewhere to act in an environment it has not yet fully observed.
Online evidence is therefore indispensable, because no other evidence about deployment-specific prior validity exists.

\textbf{The Diagnostic Dimensions.}~
The information available for making reliance decisions changes fundamentally over the course of training. Before online interaction begins, the practitioner knows only the prior and the task. 
Once online learning begins, each interaction reveals where the prior holds and where it does not. These two stages require two complementary forms of diagnosis.
 
The first form is \textit{\textbf{prior-deployment match assessment}}. Before online training begins, the practitioner should estimate how well each prior fits the task at hand, based on properties that are observable without interaction, such as data coverage, policy quality, or model accuracy~\cite{li2025three, asadulaev2025expert}.
Recent work demonstrates that even coarse estimates carry actionable information: comparing offline policy quality with data quality can already determine which component the practitioner should anchor on~\cite{li2025three}, and lightweight metrics can predict whether a dataset will support effective fine-tuning~\cite{asadulaev2025expert}.
By informing the initial configuration, match assessment enables \textit{\textbf{flexibility}} across deployments.
 
The second form is \textit{\textbf{tension dynamics monitoring}}.
Training is not static: aspects that were initially valuable may become outdated as the agent's own experience grows, turning helpful guidance into a binding constraint.
The field already recognizes this implicitly.
Methods that anneal conservative penalties~\cite{cho2025annealing}, schedule warmup phases~\cite{zhou2025efficient}, or periodically reset the reference model~\cite{liu2025prorl} all assume that reliance should change during training, but make these adjustments on a fixed schedule rather than in response to observed learning dynamics.
Emerging work shows that measurement-driven adjustment is feasible: plasticity metrics can detect capacity loss during training~\cite{sokar2023dormant, ma2024revisiting, xu2024drm}, adaptive replay buffers can rebalance data sources based on relevance signals~\cite{song2026adaptive}, and reward-model monitors can flag proxy divergence~\cite{miao2025information}.
However, these tools remain fragmented across domains, each targeting a specific failure mode rather than assessing prior validity as a whole.
By tracking how prior validity evolves during training, dynamics monitoring enables \textit{\textbf{adaptivity}} within each deployment.

In practice, these two dimensions interact. Online evidence gathered through dynamics monitoring can retrospectively validate or revise the initial match assessment, and a better initial assessment reduces the burden on runtime monitoring.
This loop is almost entirely absent today. Most methods fix their reliance configuration at the start of training or adjust it on a predetermined schedule, without using online evidence to revise the initial assessment.
Closing this loop through principled diagnostic infrastructure is, in our view, necessary for making online RL with offline priors reliably deployable.
\Takeaway{The field should shift from benchmark-driven to diagnosis-driven tension management. Online experience carries evidence not only about the task but also about the ongoing validity of each prior. Diagnosis-driven tension management is the practice of extracting this evidence and using it to guide reliance decisions, enabling both flexible and adaptive deployment.}
% \newpage
\section{Objections and Counterarguments}
\label{sec:alternative}
\vspace{-0.5\baselineskip}

% A position paper should engage with its strongest objections. 
% We consider five counter-arguments to our central claims.
We consider five potential objections to our call for diagnosis-driven online RL with offline priors.

\textit{\textbf{This is an engineering challenge, not a scientific insight.}}~
It is fair to ask whether this paper merely names a phenomenon that practitioners already navigate daily: different deployments need different configurations, and finding good ones is routine engineering.
Each community already handles its own regime dependence, but in isolation: the offline-to-online RL community studies initialization versus data retention~\cite{li2025three}, the RLHF community debates KL penalty strength~\cite{yu2025dapo, ackermann2026gradient}, and the sim-to-real community weighs fine-tuning against freezing~\cite{as2026matters}. 
Our contribution is recognizing that these are superficially different expressions of the same structural phenomenon and that the same diagnostic principles apply across all of them. 
Cross-community unification of this kind has consistently been treated as scientific contribution in ML, from the formalization of transfer learning~\cite{pan2009survey} to the systematization of catastrophic forgetting~\cite{kirkpatrick2017overcoming}.
A reader may also interpret our formalization as a hyperparameter tuning problem, but the decisions that $\mu$, $\lambda$, and $\beta$ abstract over are not points on a continuous search grid.
They abstract over architectural and algorithmic choices such as using pretrained versus random initialization, imposing versus dropping a conservative penalty, and retaining versus discarding offline data.
Furthermore, prior validity shifts as online experience accumulates, so the right reliance level at the start of training may not remain right later.

\textit{\textbf{Better offline priors solve the problem at the source.}}~
Rather than diagnosing bounded commitment during deployment, one could try to prevent it upstream. Recent work pursues this through larger and more diverse datasets~\cite{walke2023bridgedata, khazatsky2024droid}, adaptive offline objectives that preserve revision capacity~\cite{ni2026adaptive, ghosh2022offline}, world foundation models that encode broad physical priors~\cite{agarwal2025cosmos, team2026gigabrain}, and Bayesian formulations that build uncertainty directly into the learned policy~\cite{hu2024bayesian, ni2025long}.
These efforts are valuable and often dramatically improve transfer and generalization.
However, in every major deployment paradigm, online adaptation remains a necessary stage: foundation-model priors are designed as efficient initializations to be fine-tuned, not as finished policies for arbitrary deployment~\cite{kim2024openvla}, and adaptive offline objectives explicitly aim to preserve the capacity for later revision, not to remove the need for it~\cite{ni2026adaptive, ghosh2022offline}.
The reason is structural: online RL exists in the pipeline precisely because the prior does not fully solve the deployment task, and improving beyond the prior necessarily means entering territory where its guidance is no longer reliable. Better priors extend the region where reliance is safe; diagnosis addresses what happens at and beyond its boundary. The two are complementary, and advances in either make the other more effective.

\textit{\textbf{Scaling and algorithmic progress will resolve this.}}~
A longer-term version of the previous argument holds that continued progress in model scale, data scale, and algorithm design will eventually make bounded commitment negligible, rendering diagnostic infrastructure a premature investment. There is real evidence for this view: at sufficient scale, some classical pathologies weaken. Large pretrained VLAs show little forgetting during continual adaptation~\cite{hu2026simple, liu2026pretrained}, and larger language models exhibit more efficient RL post-training~\cite{tan2025scaling}. However, the strongest scaling results are themselves regime-dependent. The VLA recipe that eliminates forgetting requires a specific combination of large model, parameter-efficient tuning, and on-policy RL; removing any ingredient causes collapse~\cite{hu2026simple}. Fine-tuning capacity does not transfer uniformly across tasks and embodiments~\cite{kim2026adaptive, li2026matters}, and platform-dependent defaults persist even with state-of-the-art algorithms~\cite{as2026matters}. What scale changes is not whether deployment-specific choices matter but which ones matter most. Meanwhile, scaling expands the range of deployments the field attempts to address, introducing new embodiments, task types, and deployment conditions faster than any single advance can uniformly cover. This makes principled diagnosis more necessary as the field scales, not less.

\textit{\textbf{The field only needs better benchmarks, not diagnostic infrastructure.}}~
Recent position papers have proposed valuable reforms to RL evaluation: accounting for tuning costs~\cite{tang2025position}, restricting lifetime access~\cite{mesbahi2025position}, and testing for fragile generalization~\cite{chen2025rethinking, jordan2024position}.
This concern is well founded, and we agree that evaluation methodology needs reform. However, even perfect benchmarks answer a different question than diagnostics do.
Benchmarks tell us which method tends to work under which conditions; diagnostics tell us whether a specific deployment meets those conditions, and whether the answer is changing as training proceeds.
The second question requires online evidence that only deployment interaction can generate, which is why benchmarks and diagnostics are complementary: one narrows the space of candidate methods, the other guides their configuration during training.
% The second question requires online evidence that only the deployment interaction can generate, which is why the two are complementary rather than competing: benchmarks narrow the space of candidate methods before deployment, while diagnostics guide configuration and adjustment during it.

\textit{\textbf{Deep RL is too opaque for reliable diagnosis.}}~
Measuring prior validity is genuinely harder than measuring network capacity, and no general-purpose diagnostic toolkit exists today. However, the relevant question is not why the network behaves as it does but whether the prior is still helping. The former requires interpretability, which remains hard. The latter requires only observable quantities: performance trends, distribution overlap, and prediction accuracy~\cite{li2025three, asadulaev2025expert}. Practical building blocks already exist across domains: discrete regime distinctions guide method selection in offline-to-online RL~\cite{li2025three}, reward-quality monitors flag proxy divergence in RLHF~\cite{miao2025information}, and uncertainty estimates weight synthetic data by reliability in model-based RL~\cite{aghabozorgi2026wimle}.
Each measures observable signals and translates them into actionable decisions. 
The plasticity literature shows this trajectory is viable: from informal recognition to systematic measurement to a productive diagnostic subfield with reusable tools, all within a few years~\cite{sokar2023dormant, lyle2023understanding}.
Prior validity diagnosis is at an earlier stage of the same progression. 
Furthermore, diagnosis need not be perfect to be productive.
Even coarse measurements improve on decisions that would otherwise be made without evidence, and each deployment that uses diagnostic signals generates insights that sharpen future tools. 
This self-reinforcing cycle between diagnostics and deployment is a promising path toward reliable online RL with offline priors.
\section{Conclusion}
\label{sec:conclusion}

This paper has argued that offline priors fundamentally reshape the structure of online RL. By introducing knowledge whose scope of validity the agent cannot fully determine, priors transform the one-sided bottlenecks of from-scratch learning into genuine two-sided tensions. We have formalized this through the concept of bounded commitment, shown empirically that no single reliance configuration is universally optimal, and argued that the field should shift from benchmark-driven to diagnosis-driven tension management, in which online experience is used not only to learn the task but also to assess prior validity and guide reliance decisions toward flexible and adaptive deployment.

\textbf{Research Opportunities.}~
If the field adopts diagnosis-driven tension management, several opportunities open up that the current paradigm does not naturally support.
\begin{itemize}[leftmargin=1.5em, itemsep=1.5pt, parsep=0pt, topsep=0pt]
\item \textit{A new class of research contributions.} 
Under the benchmark-driven paradigm, contributions are measured primarily by performance gains. Diagnosis-driven research values a different kind of output: not a method that wins on a benchmark, but a signal that predicts when a design choice helps or hurts, a metric that assesses prior-deployment match, or a monitor that tracks prior validity during training~\citep{li2025three, asadulaev2025expert}.
Work of this kind already exists but is typically framed as supporting analysis rather than a primary contribution~\cite{ma2026makes, obando-ceron2026simplicial, han2026fire}.
\item \textit{Cross-community knowledge transfer.} Currently, each community rediscovers similar failure modes in isolation: catastrophic forgetting in offline-to-online RL, reward hacking in RLHF, reality-gap collapse in sim-to-real transfer. 
If these are recognized as manifestations of the same structural phenomenon, diagnostic tools developed in one community can inform practice in others. 
Plasticity metrics illustrate this potential: dormant neuron ratios originated in a specific experimental setting~\citep{sokar2023dormant} but now serve as reusable diagnostics across tasks and algorithms.
\item \textit{Deployment as a source of scientific knowledge.} Under the current paradigm, deployment is the endpoint of research: methods are developed, evaluated, and then applied. Diagnosis-driven deployment inverts this relationship. Every deployment that uses diagnostic evidence generates insights about the conditions under which each prior holds or fails, feeding back into the design of better tools and more informed future deployments.
\item \textit{Methods aware of their own bounded commitments.} The diagnosis-driven paradigm also changes how methods themselves are designed. Rather than seeking algorithms that perform well across the board, researchers can design methods that are explicitly aware of the boundaries of their prior knowledge and capable of adjusting their own reliance as those boundaries are revealed during training. This represents a shift from optimizing for average-case performance to building in the capacity for deployment-specific adaptation.
\end{itemize}

\textbf{A Broader Perspective.}~
Underlying these opportunities is a more fundamental question: \textit{what kind of knowledge should the field be accumulating?}
Under the benchmark-driven paradigm, the field accumulates methods: each validated under particular conditions, each adding to a growing catalogue that transfers poorly across deployments. 
The diagnosis-driven paradigm instead accumulates understanding: not which method wins where, but what determines when each approach works and why. Each insight about the conditions of success informs not only current practice but future method design. 
Individual methods will be superseded. Understanding of the fundamental structure of learning persists, compounds, and shapes whatever comes next.

% ── References ──────────────────────────────────────────────
\clearpage
\bibliography{ref}               % BibTeX file: ref.bib

% ── Appendix ────────────────────────────────────────────────
\newpage
\appendix
\onecolumn
\addtocontents{toc}{\protect\setcounter{tocdepth}{-1}}
% \newpage
\section{Illustrative Experiments: Full Results}
\label{app:experiments}

\Section~\ref{sec:no-universal} presents representative task pairs to illustrate the non-universality of tension management. This appendix provides the complete experimental setup and full results across all tasks and reliance dimensions.

\paragraph{Setup.}
All experiments are conducted in the offline-to-online RL setting on D4RL benchmarks~\cite{fu2020d4rl}, covering Adroit manipulation (Pen, Relocate, Door), Kitchen (Complete, Partial, Mixed), AntMaze navigation (Large-Diverse, Large-Play, Ultra-Diverse), and MuJoCo locomotion (HalfCheetah, Hopper, Walker2D, each with Random, Medium-Replay, Medium, and Medium-Expert datasets). Each experiment isolates one reliance parameter from Table~\ref{tab:structural} by toggling it while holding the others fixed. We organize the experiments into three groups corresponding to the three functional roles, with two groups further split to reveal interactions between parameters.
Table~\ref{tab:exp-conditions} summarizes the experimental conditions. In all cases, the ``with'' and ``without'' conditions differ in exactly one reliance dimension, enabling controlled comparison.

\begin{table}[ht]
\centering
\caption{Summary of experimental conditions. Each row isolates one reliance parameter while holding the others fixed. Checkmarks indicate active components.}
\label{tab:exp-conditions}
\small
\begin{tabular}{@{}llccc@{}}
\toprule
\textbf{Experiment} & \textbf{Comparison} & \textbf{Init ($\mu$)} & \textbf{Ref ($\lambda$)} & \textbf{Aux ($\beta$)} \\
\midrule
Initialization & w/ init vs w/o init & \textbf{varies} & $\times$ & \checkmark \\
\midrule
Reference (no aux) & w/ ref vs w/o ref & \checkmark & \textbf{varies} & $\times$ \\
Reference (with aux) & w/ ref vs w/o ref & \checkmark & \textbf{varies} & \checkmark \\
\midrule
Auxiliary (no ref) & w/ aux vs w/o aux & \checkmark & $\times$ & \textbf{varies} \\
Auxiliary (with ref) & w/ aux vs w/o aux & \checkmark & \checkmark & \textbf{varies} \\
\bottomrule
\end{tabular}
\end{table}

\newpage
\textbf{Initialization ($\mu$)}~
This experiment tests whether initializing the online learner from the offline-trained policy and value function improves over random initialization. Both conditions retain the offline dataset in the replay buffer following the RLPD protocol~\cite{ball2023efficient} and use standard SAC updates without conservative penalties. The comparison thus isolates the effect of initialization reliance: starting from prior parameters ($\mu = 1$) versus starting from random parameters ($\mu = 0$), with auxiliary support held constant.
\begin{figure}[ht]
\centering
\includegraphics[width=0.78\textwidth]{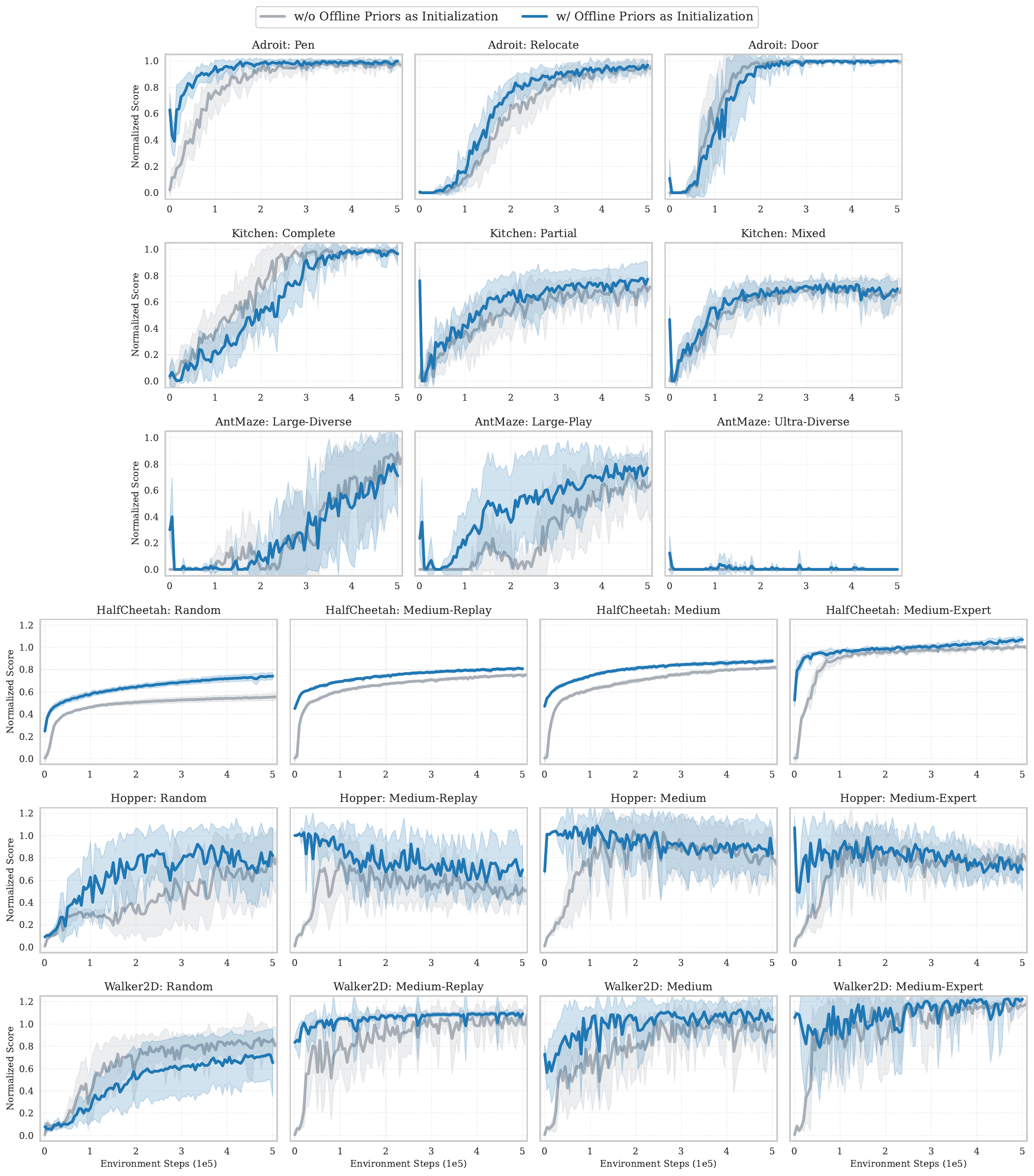}
\caption{Effect of initialization reliance ($\mu$), with auxiliary data retained in both conditions. Using offline-trained initialization helps on some tasks but hurts on others, illustrating that the optimal $\mu$ is task-dependent.}
\label{fig:init-full}
\end{figure}

\newpage
\textbf{Reference ($\lambda$), without auxiliary data.}~
This experiment tests whether maintaining a conservative value penalty during online learning (Cal-QL~\cite{nakamoto2023cal}) improves over unconstrained SAC~\cite{haarnoja2018soft}. Both conditions start from the offline-trained initialization and do not retain offline data in the replay buffer. The comparison isolates the effect of reference reliance: conservative penalty active ($\lambda > 0$) versus no penalty ($\lambda = 0$), without auxiliary support.
\begin{figure}[ht]
\centering
\includegraphics[width=0.8\textwidth]{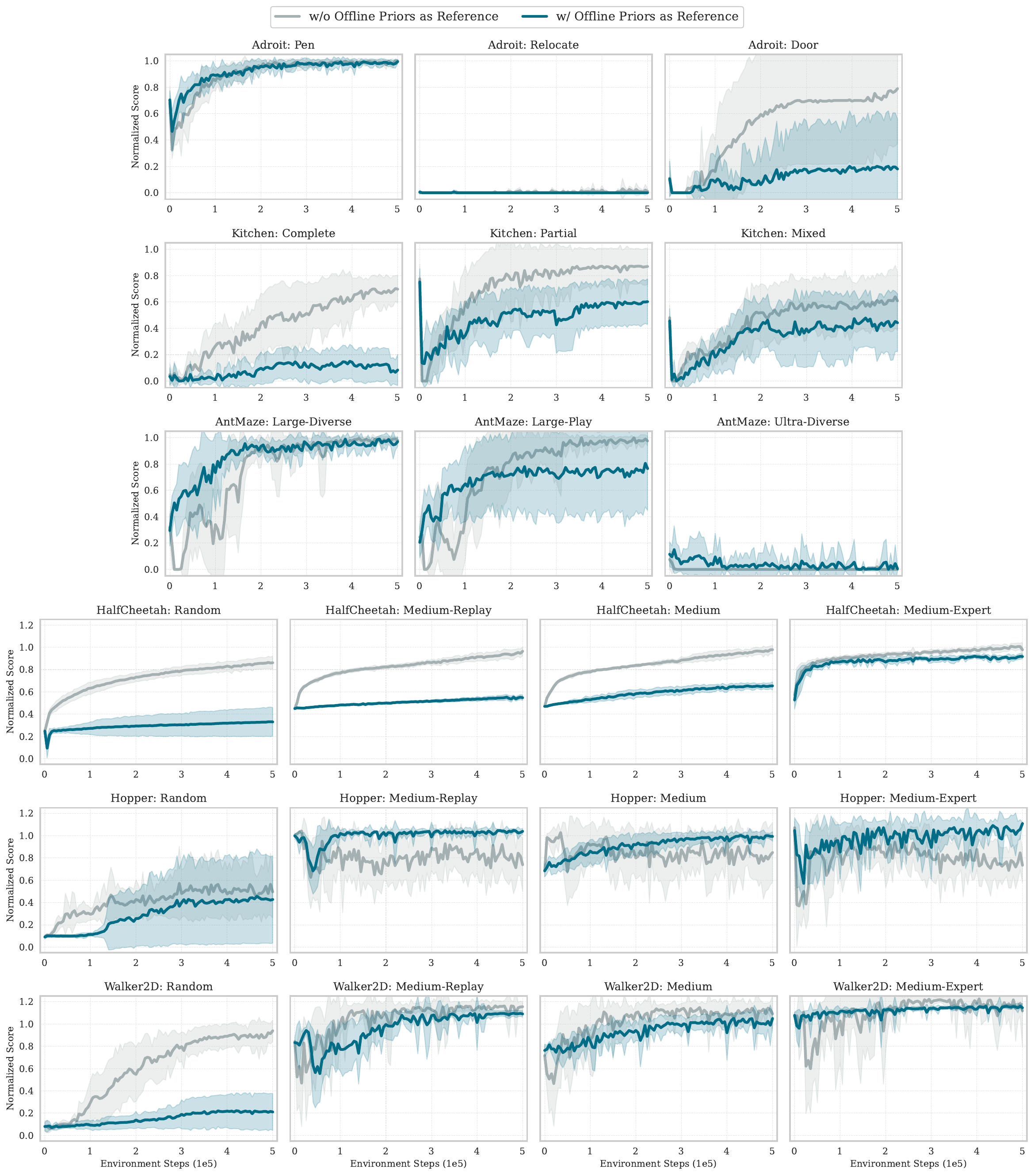}
\caption{Effect of reference reliance ($\lambda$) without auxiliary data. Both conditions use offline-trained initialization. Maintaining conservative value penalties helps on some tasks but hurts on others.}
\label{fig:ref-baseline-full}
\end{figure}

\begin{figure}[ht]
\centering
\includegraphics[width=0.75\textwidth]{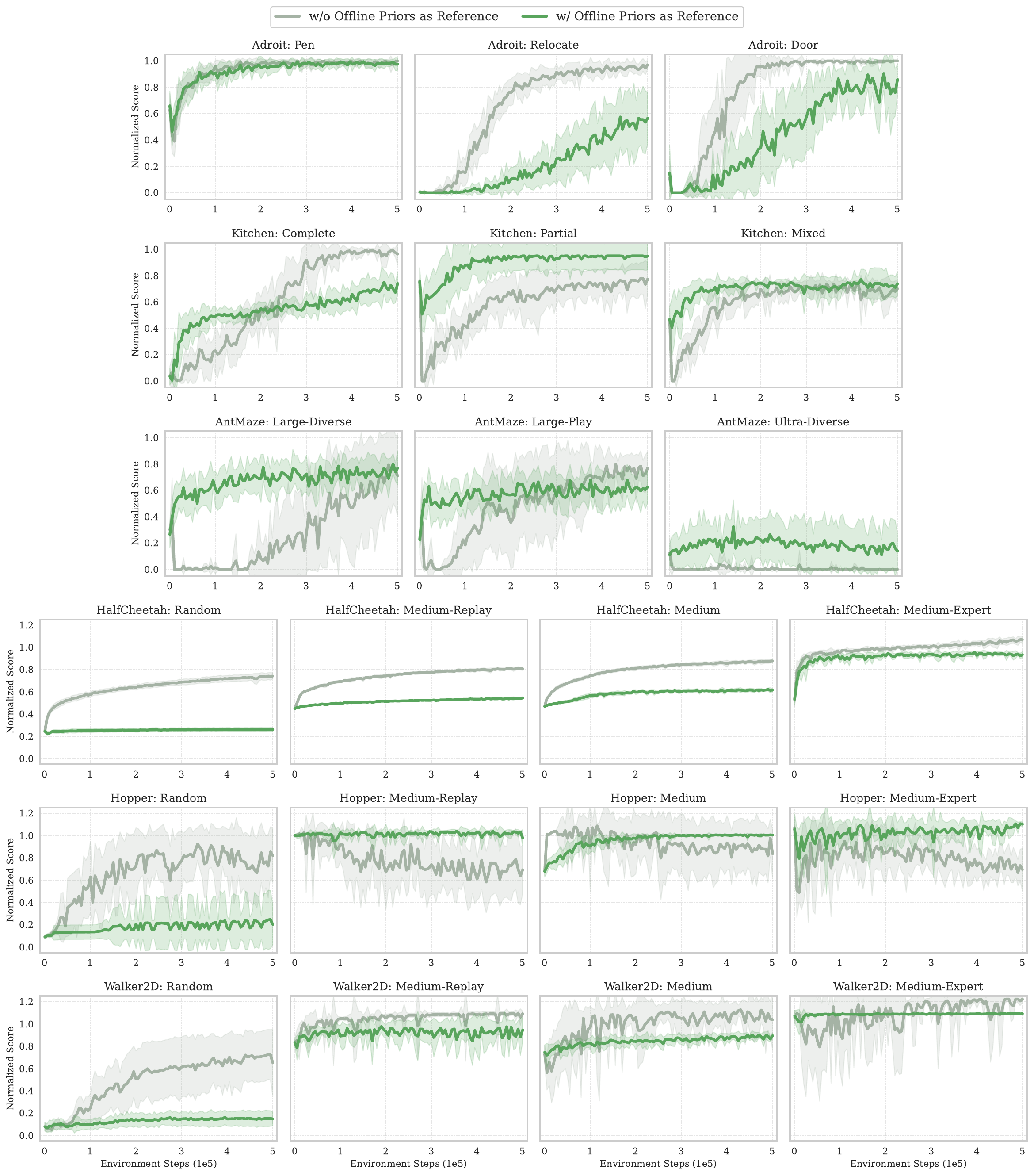}
\caption{Effect of reference reliance ($\lambda$) with auxiliary data ($\beta > 0$). Both conditions use offline-trained initialization and retain offline data. The same conservative penalty that helps or hurts in Figure~\ref{fig:ref-baseline-full} behave differently when offline data is also retained, demonstrating interaction between reliance parameters.}
\label{fig:ref-rlpd-full}
\end{figure}

\newpage
\textbf{Reference ($\lambda$), with auxiliary data.}~
This experiment repeats the reference comparison with auxiliary data present: the offline dataset is retained in the replay buffer in both conditions. The comparison isolates whether using Cal-QL's conservative penalty (as opposed to standard SAC updates) remains beneficial when offline data is also available as auxiliary support. Comparing Figures~\ref{fig:ref-baseline-full} and~\ref{fig:ref-rlpd-full} reveals how the effect of a reference constraint can itself depend on whether auxiliary data is available, illustrating the coupling between reliance parameters discussed in \Section~\ref{sec:tensions}.

% \newpage
\textbf{Auxiliary ($\beta$), without reference constraint.}~
This experiment tests whether retaining the offline dataset in the replay buffer improves over discarding it, when the online learner uses standard SAC without conservative penalties. Both conditions start from the offline-trained initialization. The comparison isolates the effect of auxiliary reliance: offline data retained ($\beta > 0$) versus online data only ($\beta = 0$), without reference constraint.
\begin{figure}[ht]
\centering
\includegraphics[width=0.8\textwidth]{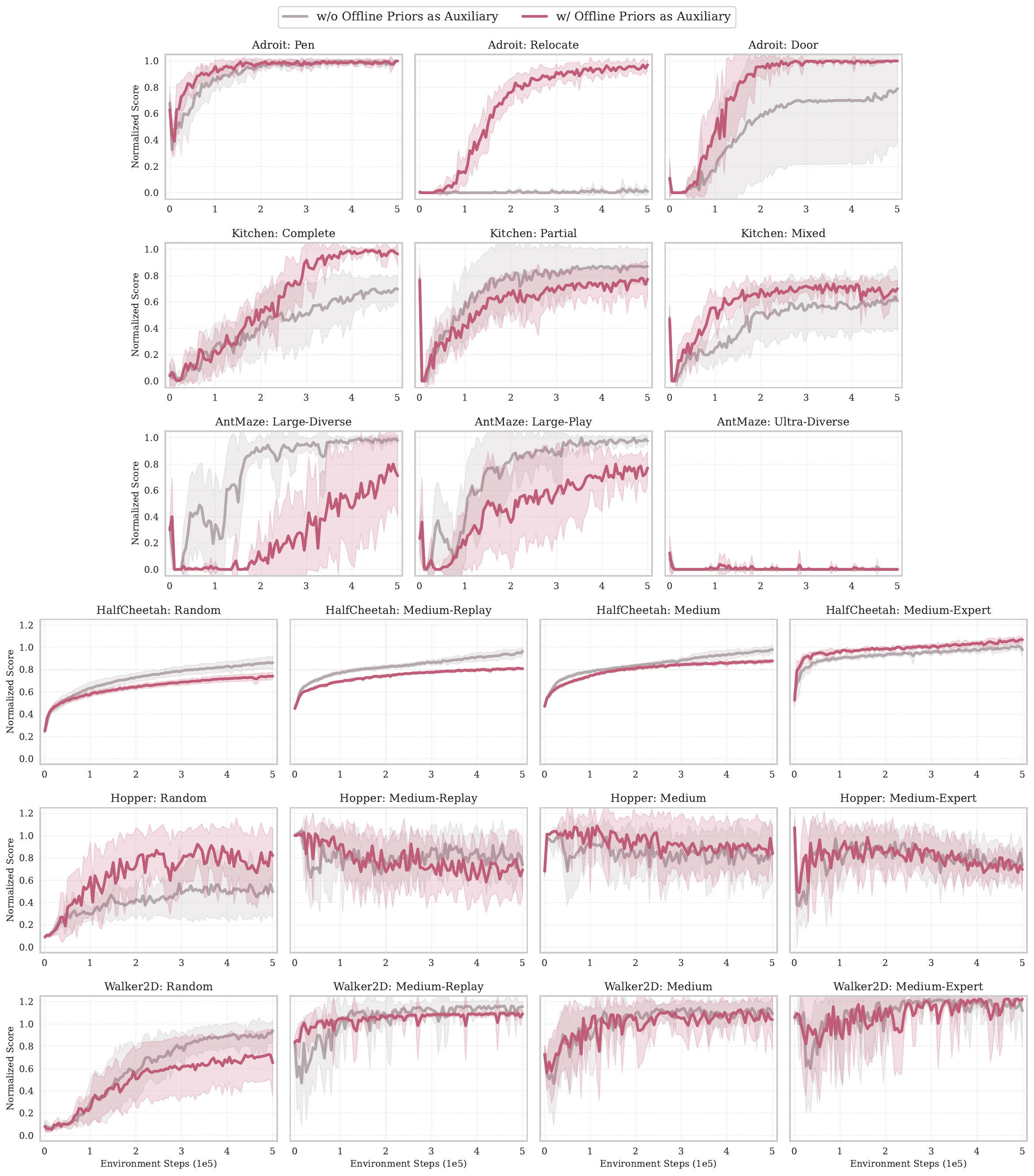}
\caption{Effect of auxiliary reliance ($\beta$) without reference constraint. Both conditions use offline-trained initialization and standard SAC. Retaining offline data helps on some tasks but hurts on others.}
\label{fig:aux-baseline-full}
\end{figure}

\begin{figure}[ht]
\centering
\includegraphics[width=0.78\textwidth]{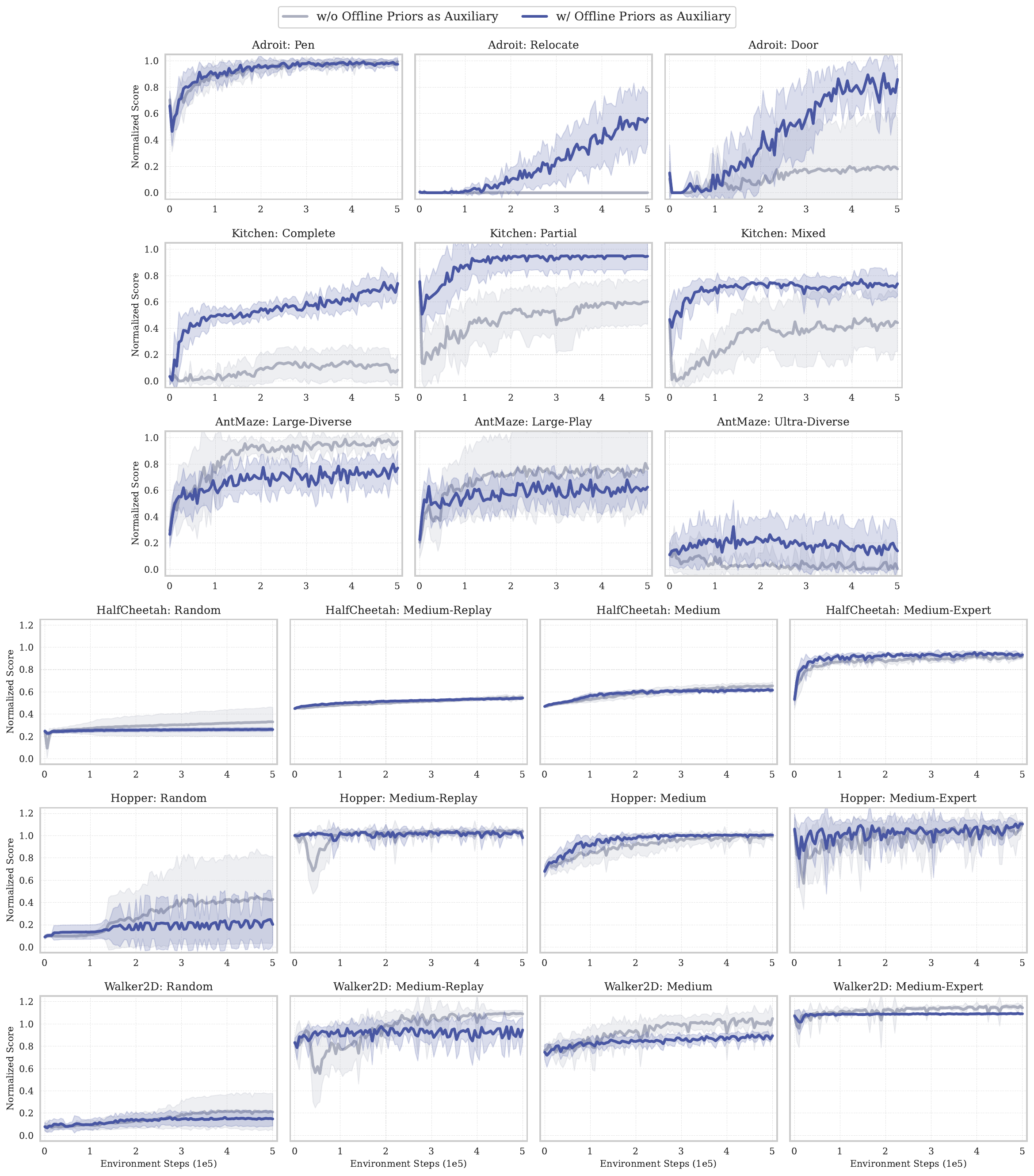}
\caption{Effect of auxiliary reliance ($\beta$) with reference constraint ($\lambda > 0$). Both conditions use offline-trained initialization and Cal-QL's conservative penalty. The effect of retaining offline data can differ from Figure~\ref{fig:aux-baseline-full}, demonstrating interaction between reliance parameters.}
\label{fig:aux-calql-full}
\end{figure}

% \newpage
\textbf{Auxiliary ($\beta$), with reference constraint.}~
This experiment repeats the auxiliary comparison with a reference constraint present: both conditions use Cal-QL's conservative penalty during online learning. The comparison isolates whether retaining offline data remains beneficial when conservative value estimation is also active. Comparing Figures~\ref{fig:aux-baseline-full} and~\ref{fig:aux-calql-full} reveals how the effect of auxiliary data depends on whether a reference constraint is in place, further confirming that the reliance parameters interact.

% \newpage
\textbf{Summary.}~
Across all five experiments and all task domains, the same pattern emerges: toggling any single reliance parameter produces help-or-hurt reversals across tasks. No setting of $\mu$, $\lambda$, or $\beta$ is uniformly beneficial. The paired experiments further reveal that the effect of one parameter depends on the configuration of the others: the impact of a reference constraint differs depending on whether auxiliary data is present (Figures~\ref{fig:ref-baseline-full} vs~\ref{fig:ref-rlpd-full}), and the impact of auxiliary data differs depending on whether a reference constraint is active (Figures~\ref{fig:aux-baseline-full} vs~\ref{fig:aux-calql-full}). This confirms that the reliance parameters are not independently tunable and supports the argument in Section~\ref{sec:no-universal} that tension management has no universal optimum.

\end{document}